\documentclass[10pt,twocolumn,letterpaper]{article}

\usepackage[pagenumbers]{iccv} 

\usepackage{amsmath}        
\usepackage{amssymb}        
\usepackage{algorithm}      
\usepackage{algpseudocode}  
\usepackage{graphicx}       
\usepackage{xcolor}         
\usepackage{multirow}
\usepackage{makecell}
\usepackage{colortbl}

\definecolor{iccvblue}{rgb}{0.21,0.49,0.74}
\usepackage[pagebackref,breaklinks,colorlinks,allcolors=iccvblue]{hyperref}
\def\paperID{11666} 
\def\confName{ICCV}
\def\confYear{2025}

\title{Training-free Diffusion Acceleration with Bottleneck Sampling}

\author{ Xin Xia$^{2}$\quad Ye Tian$^{1^*}$\thanks{Equal Contribution.} \quad Yuxi Ren$^{2}$ \quad  Shanchuan Lin $^{2}$\\ 
Xing Wang$^{2}$ \quad Xuefeng Xiao$^{2}$\thanks{Project Leader.} \quad  Yunhai Tong$^{1}$ \quad Ling Yang$^{1}$\thanks{Corresponding Authors.} \quad  Bin Cui$^{1 \ddag }$
\and
$^1$Peking University \quad $^2$Bytedance  \\
Project: \href{https://tyfeld.github.io/BottleneckSampling.github.io/}{Bottleneck-Sampling-Page}
}

\begin{document}
\maketitle

\begin{abstract}
Diffusion models have demonstrated remarkable capabilities in visual content generation but remain challenging to deploy due to their high computational cost during inference. This computational burden primarily arises from the quadratic complexity of self-attention with respect to image or video resolution. While existing acceleration methods often compromise output quality or necessitate costly retraining, we observe that most diffusion models are pre-trained at lower resolutions, presenting an opportunity to exploit these low-resolution priors for more efficient inference without degrading performance. In this work, we introduce \textbf{Bottleneck Sampling}, a training-free framework that leverages low-resolution priors to reduce computational overhead while preserving output fidelity. Bottleneck Sampling follows a high-low-high denoising workflow: it performs high-resolution denoising in the initial and final stages while operating at lower resolutions in intermediate steps. To mitigate aliasing and blurring artifacts, we further refine the resolution transition points and adaptively shift the denoising timesteps at each stage. We evaluate Bottleneck Sampling on both image and video generation tasks, where extensive experiments demonstrate that it accelerates inference by up to 3$\times$ for image generation and 2.5$\times$ for video generation, all while maintaining output quality comparable to the standard full-resolution sampling process across multiple evaluation metrics. 


\end{abstract}    
\section{Introduction}
\begin{figure}[ht]
    \centering
    \includegraphics[width=1\linewidth]{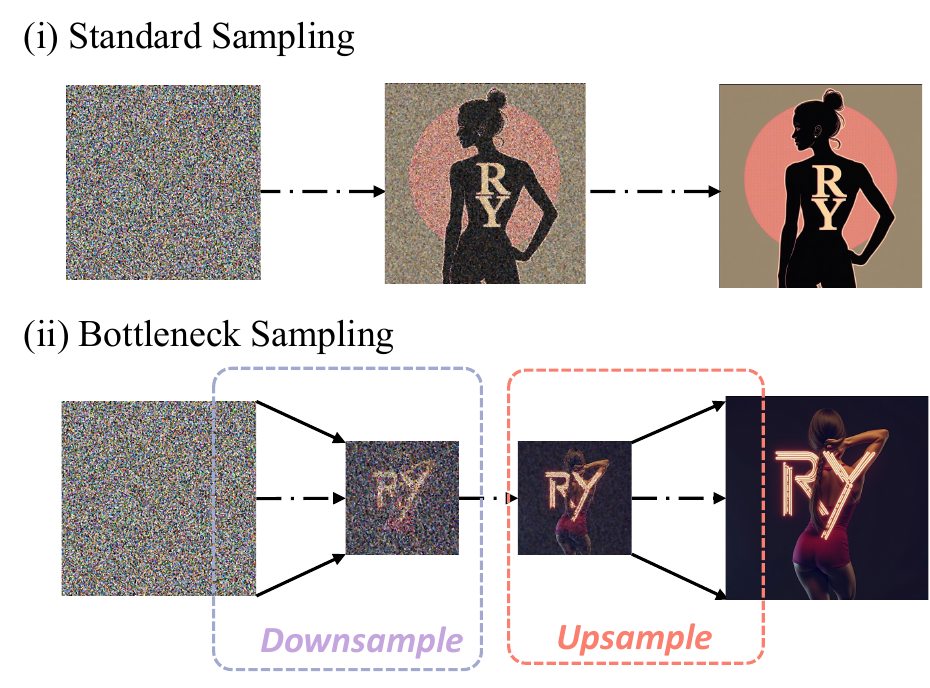}
    \caption{Comparison of sampling strategies in our framework. (i) Standard Sampling. (ii) Our Bottleneck Sampling: a high-low-high workflow that captures semantics early, improves efficiency in the middle, and restores details at the end. Images generated by FLUX.1-dev using the prompt: ``Design a stylish dancer's back logo with the letters 'R' and 'Y'".}
    \label{fig:intro}
    \vspace{-1.5em}
\end{figure}

\begin{figure*}[ht]
    \centering
    \includegraphics[width=0.98\linewidth]{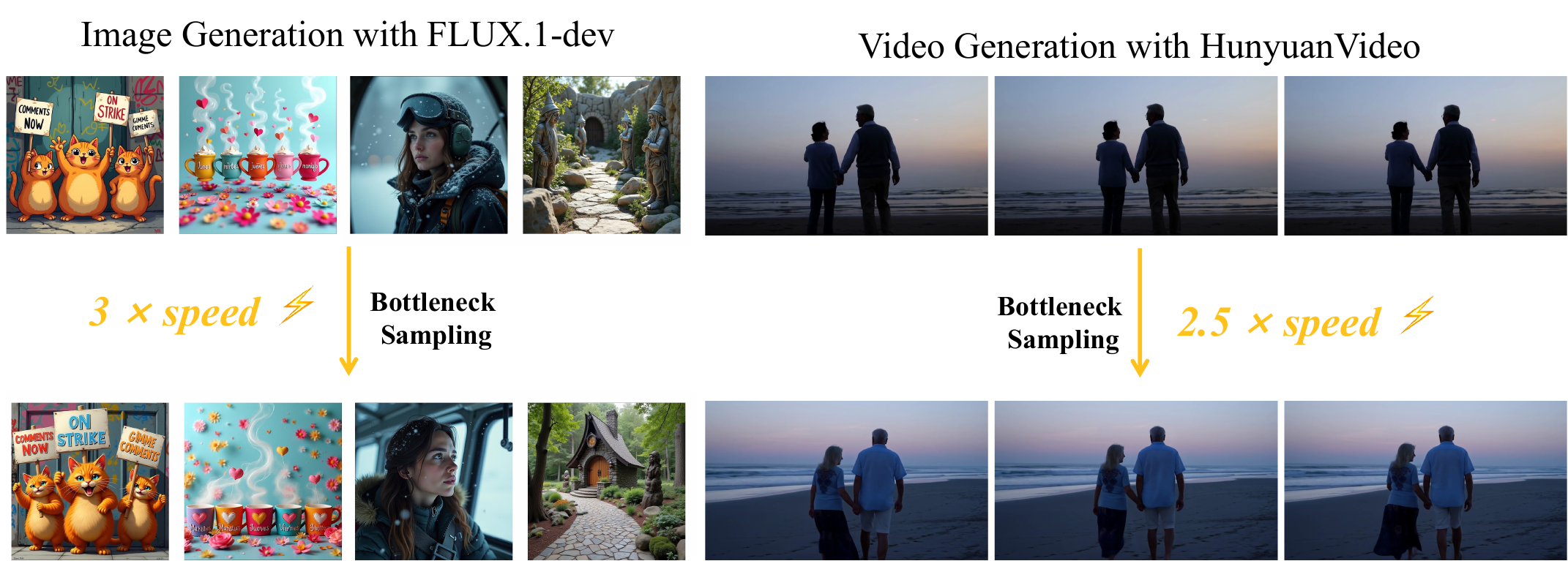}
    \caption{Main Results of our Bottleneck Sampling on both text-to-image generation and text-to-video generation. Bottleneck Sampling maintains comparable performance with a 2.5 - 3 $\times$ acceleration ratio in a training-free manner.}
    \label{fig:main_results}
     \vspace{-1em}
\end{figure*}

Denoising Diffusion models have emerged as a cornerstone of modern generative modeling, achieving state-of-the-art performance in tasks such as text-to-image synthesis and video generation~\citep{ho2020denoising,song2020score,sohl2015deep, ramesh2022hierarchical, rombach2022high, ho2022imagen, yang2024mastering, zhang2024itercomp}. Among the recent developments, Diffusion Transformers (DiTs)~\citep{peebles2023scalable} have gained significant attention due to their superior scalability, establishing them as the dominant approach in both image \citep{chen2023pixart, esser2024scaling, flux} and video generation \citep{yang2024cogvideox, kong2025hunyuanvideosystematicframeworklarge, tian2024videotetris}. Despite their impressive performance, the practical deployment of DiTs is hindered by their considerable computational cost, particularly during inference. The major bottleneck in the efficiency of dit is the attention's quadratic complexity of \( \mathcal{O}(L^2) \) with respect to the input token length $L$ that are determind by the image/video resolution. As the resolution of the generated images or videos increases, this quadratic scaling leads to prohibitive computational costs. For instance, generating a 2K image~\citep{chen2024pixart} that is tokenized into 16k tokens results in several seconds of attention computation even on high-end Nvidia A100 GPU. This problem becomes particularly severe when generating high-resolution content, where $L$ increases significantly, causing inference to be both time-consuming and resource-intensive.

Existing efforts to address this issue focus on optimizing attention computation within the high-resolution regime. Techniques such as attention optimization~\citep{yuan2025ditfastattn, han2024agent} and feature reuse~\citep{ma2024deepcache, ma2024learning} aim to reduce computational redundancy by discarding less important tokens or reusing intermediate activations. While these methods achieve some speedup, they inevitably introduce trade-offs in generation quality, as aggressive sparsification or reuse can degrade fine-grained details. Moreover, such approaches often require extensive hyperparameter tuning to balance efficiency and performance, making them less practical for widespread adoption. Crucially, these approaches overlook a fundamental property of corrent SOTA diffusion models: \textbf{their pretraining at lower resolutions~\citep{podell2023sdxl, yang2024cogvideox} (e.g., 256$\times$256)}, where sequence lengths are orders of magnitude smaller, and attention computations remain tractable.

\textit{This observation raises a pivotal question:} \textbf{Can we leverage low-resolution pretrained priors to accelerate high-resolution inference without sacrificing fidelity?} We hypothesize that strategically incorporating low-resolution computations during inference can substantially reduce computational overhead while preserving high-resolution output fidelity. Previous studies have explored a similar concept in cascaded diffusion models~\citep{ho2022cascaded, jin2024pyramidal}, where a low-resolution image undergoes denoising before being upsampled for super-resolution generation. However, when directly applied to inference, these methods often result in fine detail loss or text degradation due to the absence of high-resolution priors.  

To address these limitations, we introduce \textbf{Bottleneck Sampling}, a novel, training-free framework that redefines diffusion inference through hierarchical resolution scheduling. Our key insight is that \textit{early denoising stages primarily establish global structure}—a process efficiently handled at lower resolutions—while \textit{later stages refine high-frequency details} at full resolution. To achieve this, we propose a bottleneck-inspired sampling strategy that first compresses information, processes it efficiently at reduced resolutions, and then progressively restores full resolution to recover fine details. Specifically, Bottleneck Sampling begins with high-resolution noise initialization, transitions to lower resolutions for efficient global denoising, and gradually reinstates resolution for detail refinement. To mitigate aliasing and blurring artifacts, we further optimize resolution changing points and adjust denoising timesteps at each stage.  

We apply Bottleneck Sampling to both text-to-image and text-to-video models, demonstrating its versatility across different generative tasks.
Extensive evaluations across diverse metrics show that Bottleneck Sampling achieves up to 3$\times$ speedup on FLUX.1-dev\cite{flux} for image generation and 2$\times$ on hunyuanvideo\citep{kong2025hunyuanvideosystematicframeworklarge} for video generation, all while maintaining comparable output quality. Notably, our method requires no architectural modifications or retraining, making it a plug-and-play acceleration strategy for existing diffusion frameworks. By effectively bridging the efficiency-quality trade-off in high-resolution generation, Bottleneck Sampling enhances the practical deployment of diffusion models in resource-constrained environments.

\section{Related Work}

\paragraph{Diffusion Models}
Diffusion models have established themselves as a dominant paradigm in generative modeling, surpassing the performance of traditional generative adversarial networks. Early implementations, such as those by~\citet{ho2020denoising} and~\citet{rombach2022high}, utilized U-Net architectures for iterative denoising. To address scalability limitations, recent advancements have transitioned to transformer-based architectures. Notably, the Diffusion Transformer (DiT)~\citep{peebles2023scalable, chen2023pixart} replaces U-Net with a transformer backbone, achieving superior scalability in high-resolution image and video generation tasks. State-of-the-art text-to-image generation models, such as SD3~\citep{esser2024scaling} and Flux~\citep{flux}, as well as video generation models like CogVideo~\citep{yang2024cogvideox} and HunyuanVideo~\citep{kong2025hunyuanvideosystematicframeworklarge},  share a critical design principle: initial pretraining at low resolutions (e.g., $256\times256$) followed by resolution-specific fine-tuning. While effective, these approaches inherit the quadratic computational complexity of self-attention mechanisms, which becomes prohibitive at high resolutions.  In this work, we for the first time explore a novel paradigm that leverages low-resolution pretrained priors to enable efficient high-resolution generation. 

\paragraph{Image Pyramid}  
The concept of image pyramids has long been a fundamental tool in computer vision, facilitating multi-scale analysis, efficient visual data processing, and various visual understanding tasks~\citep{adelson1984pyramid, lin2017feature, wang2020deep}. More recently, image pyramids have been explored in generative models, particularly in cascaded diffusion frameworks, where generation is first performed at a low resolution and subsequently refined via super-resolution~\citep{ho2022cascaded, saharia2022photorealistic, pernias2023wurstchen, teng2023relay}. 
More recent studies have investigated the use of pyramid sampling to enhance generation efficiency. Methods such as Efficient Diffusion Transformers~\citep{chen2025edt} and Pyramid Flow Matching~\citep{jin2024pyramidal} have demonstrated the potential of coarse-to-fine generation pipelines in both image and video synthesis by restructuring network architectures and training paradigms. While these methods achieve promising results in terms of both efficiency and quality, they require extensive computational resources for retraining, making their adoption computationally expensive.  
In contrast, our method leverages low-resolution pretrained priors to propose a novel \textbf{Bottleneck Sampling} framework. Unlike existing approaches, Bottleneck Sampling requires no additional training, achieving comparable performance with minimal computational overhead.

\paragraph{Training-Free Diffusion Acceleration}  
To improve the efficiency of diffusion models, numerous training-free acceleration techniques have been proposed, primarily focusing on feature reuse and attention optimization. Feature reuse methods leverage caching to eliminate redundant computations without requiring retraining. For instance, DeepCache~\citep{ma2024deepcache} accelerates Stable Diffusion~\citep{rombach2022high} by reusing intermediate U-Net features, while Faster Diffusion~\citep{li2023faster} caches encoder features across timesteps to skip redundant computations. Learning-to-Cache~\citep{ma2024learning} further introduces an adaptive caching policy, and ToCa~\citep{zou2024accelerating} extends this idea by caching token-wise attention features in DiT for state-of-the-art acceleration. In video generation, Pyramid Attention Broadcast~\citep{zhao2024real} enables multi-dimensional attention reuse, significantly improving multi-frame synthesis efficiency.
Beyond caching, attention optimization methods directly reduce computational redundancy. Agent Attention~\citep{han2024agent} condenses information using proxy tokens, lowering the cost of full self-attention while maintaining critical dependencies. Similarly, DiTFastAttn~\citep{yuan2025ditfastattn} employs a modified window attention mechanism to focus computation on essential queries and keys.
While these approaches achieve notable speedups, they inherently introduce trade-offs. Feature reuse disrupts alignment between training and inference, potentially affecting generation quality, whereas attention optimization, despite reducing computational overhead, may lead to information loss due to selective attention reduction.

\section{Method}

\begin{figure*}
    \centering
    \includegraphics[width=1\linewidth]{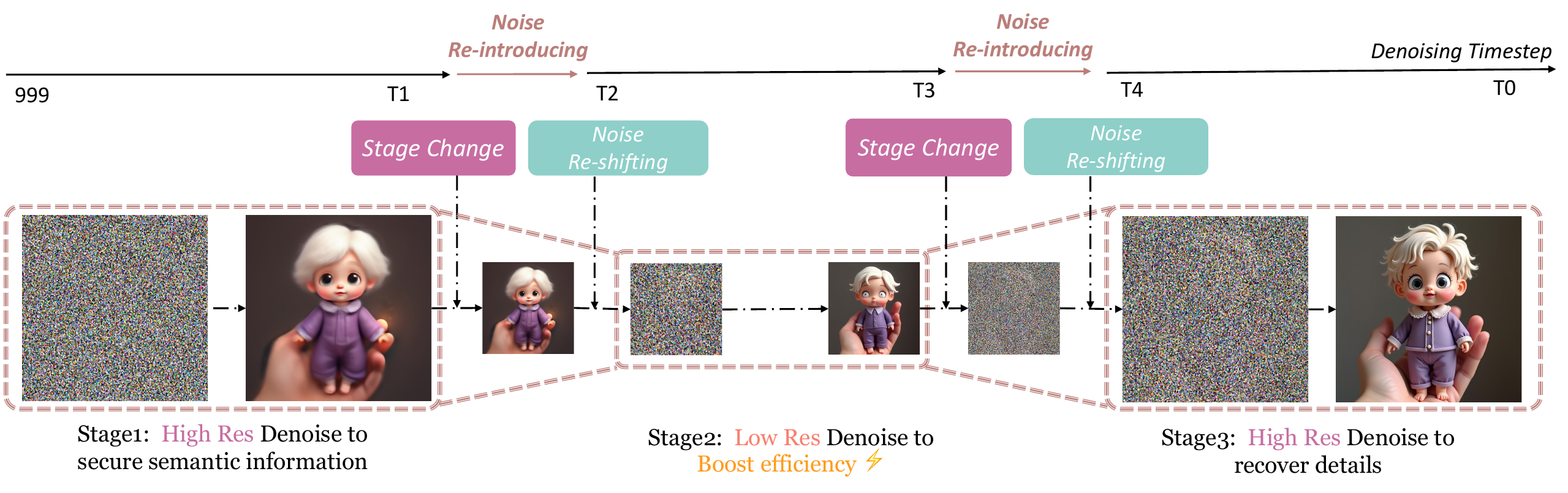}
    \caption{Overall pipeline of our Bottleneck Sampling. The process consists of three stages: (i) High-Resolution Denoising to preserve semantic information, (ii) Low-Resolution Denoising to improve efficiency, and (iii) High-Resolution Denoising to restore fine details. Images generated by FLUX.1-dev using the prompt: \textit{``2D cartoon,Diagonal composition, Medium close-up, a whole body of a classical doll being held by a hand, the doll of a young boy with white hair dressed in purple, He has pale skin and white eyes.".}}
    \label{fig:method}
\end{figure*}

\subsection{Preliminaries}
Modern state-of-the-art generative models, such as Flux~\citep{flux} and HunyuanVideo~\citep{kong2025hunyuanvideosystematicframeworklarge}, have widely adopted Flow Matching~\citep{esser2024scaling, liu2022flow, lipman2022flow, yang2024consistency, wang2024rectified} as a preferred alternative to traditional Denoising Diffusion Probabilistic Models (DDPM) due to its superior training efficiency and faster convergence. Flow Matching reformulates generative modeling as learning a continuous transformation between probability distributions \citep{yang2024consistency, wang2024rectified}, mapping complex data distributions to a simple prior (e.g., Gaussian) through a learned velocity field, and enabling sample generation via inverse integration.

During training, given a data sample \( X_1 \) and Gaussian noise \( X_0 \sim \mathcal{N}(0, \mathbf{I}) \), we define a linear interpolation trajectory:

\begin{align}
X_t = (1 - t) X_0 + t X_1, \quad t \in [0,1],
\end{align}

where \( X_t \) represents the interpolated state at timestep \( t \). The ground-truth velocity field is given by:

\begin{align}
V_t = \frac{dX_t}{dt} = X_1 - X_0,
\end{align}

capturing the instantaneous transformation from noise to data. The model \( u_\theta(X_t, y, t) \), conditioned on optional inputs \( y \) (e.g., text prompts), is optimized to minimize the velocity alignment loss:

\begin{align}
\mathcal{L} = \mathbb{E}_{t,X_0,X_1,y} \left[ \| u_\theta(X_t, y, t) - V_t \|^2 \right],
\end{align}

where \( t \) is sampled uniformly. This objective ensures the model learns a direct transport path, improving efficiency and sample quality.

At inference, starting from random noise \( X_0 \sim \mathcal{N}(0, \mathbf{I}) \), we reconstruct \( X_1 \) via ODE-based integration over \( n \) discretized timesteps \( \{t_i\}_{i=0}^n \):

\begin{align}
X_1 = X_0 + \sum_{i=0}^{n-1} u_\theta(X_{t_i}, y, t_i) \cdot (t_{i+1} - t_i),
\end{align}

where \( t_0 = 0 \) and \( t_n = 1 \). The learned velocity field \( u_\theta \) iteratively refines \( X_t \), enabling high-quality synthesis with minimal sampling steps.

\subsection{Bottleneck Sampling}


In this section, we detail the proposed Bottleneck Sampling pipeline. Taking image generation as an example, consider generating an image with height $h$ and width $w$,  during the generation phase, we start from an initial noise sample \( x_0 \in \mathbb{R}^{b \times c \times h \times w} \), where \( b \) is the batch size and \( c \) is the number of channels. Our objective is to generate a predicted output \( x_1 \) by progressively refining this initial noise. 
To achieve efficient inference while preserving quality, we design a low-high-low sampling pipeline. Specifically, during the intermediate denoising steps, we perform inference at progressively lower resolutions to reduce computational overhead. In the later stages, we gradually restore the resolution, refining the details at full scale.

Specifically, let there be \( K \) stages, characterized by a hierarchical resolution schedule satisfying \( h_1 > h_2 > \dots > h_{K-1} < h_K \), with corresponding widths \( w_1 > w_2 > \dots > w_{K-1} < w_K \). We predefine the number of inference steps in each stage as \( N_1, N_2, \dots, N_K \), where the total inference steps sum to \( T = \sum_{i=1}^{K} T_i \). At each stage \( i \), we construct discretized timesteps $\{ t_{i,j} \}_{i=0}^{K}, \quad j = 0, 1, \dots, N_i. $, the inference process follows:
\begin{equation}
\label{denoise}
X_{t_{i,j}} = X_{t_{i,j-1}} + u_\theta(X_{t_{i,j-1}}, y, t_{i,j-1}) \cdot (t_{i,j} - t_{i,j-1})
\end{equation}

where $j \in \{0, 1, \dots, N_i\}$.
\paragraph{Resolution Change Point}
After establishing the overall workflow, the next critical aspect is the resolution change.  Conventional methods~\citep{jin2024pyramidal} adjust the mean and variance of noise at boundaries to ensure smooth transitions and reduce error propagation. However, we find a more effective approach: rather than directly connecting resolution stages, we reintroduce noise to intermediate latents, enabling a fresh high-to-low noise denoising process at the new resolution.
This strategy offers two advantages. First, it aligns inference with training distribution, avoiding mismatches caused by direct latent connections. Second, it leverages the model’s multi-resolution priors. Since diffusion models learn to denoise latents based on given conditions, adding noise at transitions enables natural refinement at the new resolution, ensuring a more coherent and efficient generation process.

Specifically, at each stage transition, we first apply a standard upsampling or downsampling operation to adjust the resolution. The transition is defined as:

\begin{equation}
\label{eq:stage_change}
X_{t_{i, 0}} =
\begin{cases}
\mathrm{Up}(X_{t_{i-1, N_{i-1}}}, h_{i}, w_{i}), & \text{if } h_{i} > h_{i-1} \\
\mathrm{Down}(X_{t_{i-1, N_{i-1}}}, h_{i}, w_{i}), & \text{if } h_{i} < h_{i-1}
\end{cases}
\end{equation}

where \( X_{t_{i-1,N_{i-1}}} \) represents the final latent at stage \( i-1 \), which is then upsampled or downsampled to match the resolution of stage \( i \).

Following resolution adjustment, we reintroduce noise through Flow Matching's noise addition mechanism and subsequently perform denoising for \( t_i \) steps. The noise injection strength at each stage is parameterized by \(\{w_i\}_{i=0}^{n}\), representing distinct noise levels across different stages. The target timestep for noise injection and the noise addition are mathematically defined as:

\begin{align}
    \tau_i &= t_{i, N_i(1 - w_i)} \\
    \tilde{X}_{t_{i, 0}} &= (1 -\tau_i) X_{t_{i, 0}} +\tau_i \eta, \quad \eta \sim \mathcal{N}(0, I)
\end{align}

This formulation can be interpreted as applying noise to the latent representations from the preceding stage with a controlled intensity of \(w_i\). As a result, the subsequent stage requires only \(N_i w_i\) steps rather than the complete \(N_i\) steps. Our experimental results demonstrate that this architectural design steadily enhances the stability and efficiency of both image and video generation processes. 


\paragraph{Tailored Scheduler Re-Shifting}

\begin{figure}[ht]
    \centering
    \includegraphics[width=0.95\linewidth]{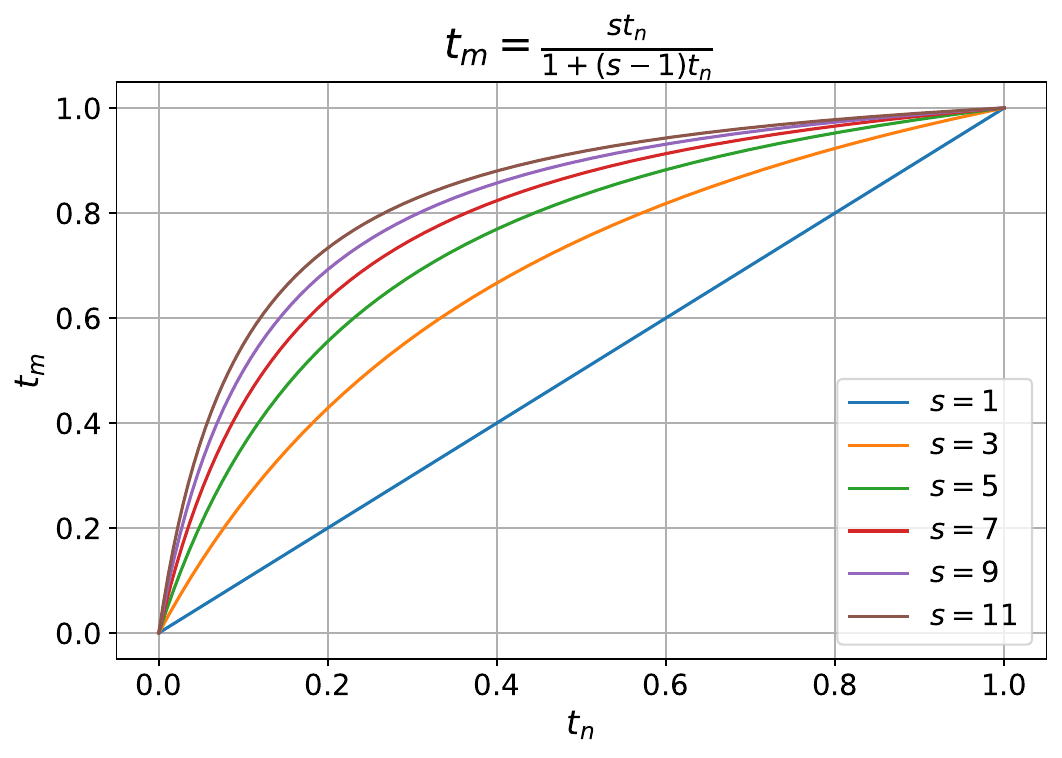}
    \caption{Timestep Shifting Visualization at different shifting factors settings. Higher shifting scales lead to denoising in higher-noise regions}
    \label{fig:timetep}
    \vspace{-1em}
\end{figure}

Following the handling of stage changing, a crucial subsequent step involves adapting the denoising scheduler for the next stage to accommodate potential variations in the signal-to-noise ratio (SNR).  
In contrast to DDPM~\citep{ho2020denoising}, which employs a linear scheduling strategy throughout the inference steps, flow matching typically adopts a shifting formulation~\citep{esser2024scaling, kong2025hunyuanvideosystematicframeworklarge, flux}, as illustrated in \cref{fig:timetep}. This approach emphasizes denoising in high-noise regions, aligning with prior research demonstrating that concentrating the denoising process in low-SNR regions yields superior outcomes.  
Within our framework, resolution adjustments not only modify spatial characteristics but also influence the SNR properties of each latent region, effectively diminishing the retained signal from the preceding stage. Given the reintroduction of noise during stage transitions, each latent representation inherently reverts to a low-SNR regime. To mitigate this, we incorporate an additional scheduler shift at stage transitions, facilitating more stable denoising. Let the shift factors be defined as $\{s\}^n_{i=0}$
At each stage, the scheduling process is then adjusted as:
\begin{align}
\label{eq:shifting}
t_{i,m} = \frac{s_i \cdot t_{i,n}}{1 + (s_i - 1) t_{i,n}}.
\end{align}
where $t_{i,n}$ is the original timestep and $t_{i,m}$ is the modified timestep actually used in \cref{denoise} with scale factor $s_i$. This re-shifting strategy ensures that denoising aligns with the shifting SNR across resolutions, leading to improved generative performance in both image and video generation.

\subsection{Overall Algorithm of Bottleneck Sampling}
\begin{algorithm}[H]
\caption{\textbf{Bottleneck Sampling}}
\label{alg:Bottleneck Sampling}
\begin{algorithmic}[1]
\Require Noise sample $ x_0 \in \mathbb{R}^{b \times c \times h \times w} $, sampling stages $K$, resolution list $(h_1, w_1), (h_2, w_2), \dots, (h_K, w_K) $, number of inference steps $\{N_i\}_{i=0}^K$, noise injection strengths $ \{w_i\}_{i=0}^K $, shifting factors $\{s_i\}_{i=0}^K$
\Ensure Generated output $ X_1 = X_{K, N_K} $
\State Initialize$ X_{t_{0,0}} \gets x_0 $
\For{$ i = 1, \dots, K $}
    \If{$ i > 1 $}
        \State \textcolor{blue}{//\quad Stage Change Points:}
        \State Upsample or Downsample $ X_{t_{i-1, N_{i-1}}} $ to $ X_{t_{i, 0}} $ using \cref{eq:stage_change}
        \State \textcolor{blue}{//\quad Tailoerd Scheduler Shifting:}
        \State Update flow matching scheduler with shift factor $ s_i $ using \cref{eq:shifting}
       
    \EndIf
    \State Crafting timesteps $ \{t_{i, j}\}_{j=0}^{N_i} $ using tailored scheduler after re-shifting.
     \State Adjust inference steps: $ N_i \gets (1 - w_i) \cdot N_i $
    \For{$ j = 1, \dots, N_i $}
        \State Compute velocity field $ u_\theta(X_{t_{i,j}}, y, t_{i,j}) $
        \State Perform denoising step using \cref{denoise}
    \EndFor
\EndFor
\State \Return $ X_1 \gets X_{K, N_K} $
\end{algorithmic}
\end{algorithm}

\section{Experiments}

\subsection{Settings}
\paragraph{Baselines}
We evaluate Bottleneck Sampling on two widely used diffusion transformers: FLUX.1-dev\citep{flux} for image generation and HunyuanVideo\citep{kong2025hunyuanvideosystematicframeworklarge} for video generation. Both models, based on MM-DiT\citep{esser2024scaling} with a flow matching scheduler, enable a unified implementation of our approach. We primarily compare against ToCa\citep{zou2024accelerating}, a state-of-the-art training-free acceleration method that employs token-wise caching for DiTs. Additionally, we include varients of baseline that directly reduce inference steps. 

\paragraph{Metrics}
For both text-to-image and text-to-video tasks, we use CLIP Score\citep{radford2021learning} to assess the alignment between generated content and textual prompts. For image generation, we further employ ImageReward\citep{xu2024imagereward} for human preference evaluation, along with compositional metrics GenEval\citep{ghosh2023geneval} and T2I-Compbench\citep{huang2023t2i}. For video generation, we adopt VBench\citep{huang2024vbench} to evaluate overall quality and T2V-Compbench\citep{sun2024t2v} for compositional capabilities. To account for the subjectivity in evaluation, we also conduct an extensive user study detailed in \cref{app:user_study}.

\paragraph{Stage Configurations}
We explore different stage configurations for image and video generation and report the optimal settings in our evaluations. For image generation with FLUX.1-dev, we adopt a three-stage pipeline with resolutions \(1024 \to 512 \to 1024\). For video generation, we use the same sampling strategy while adjusting the resolution to fit the constraints of HunyuanVideo, adopting a \(1240p \to 738p \to 1240p\) configuration. Lanczos resampling~\citep{duchon1979lanczos} is used for both upsampling and downsampling. Detailed hyperparameters and ablation studies are provided in \cref{app:hyp} and \cref{app:ablation}.  

\subsection{Results on Image Generation}
\begin{figure*}[ht]
    \centering
    \includegraphics[width=1\linewidth]{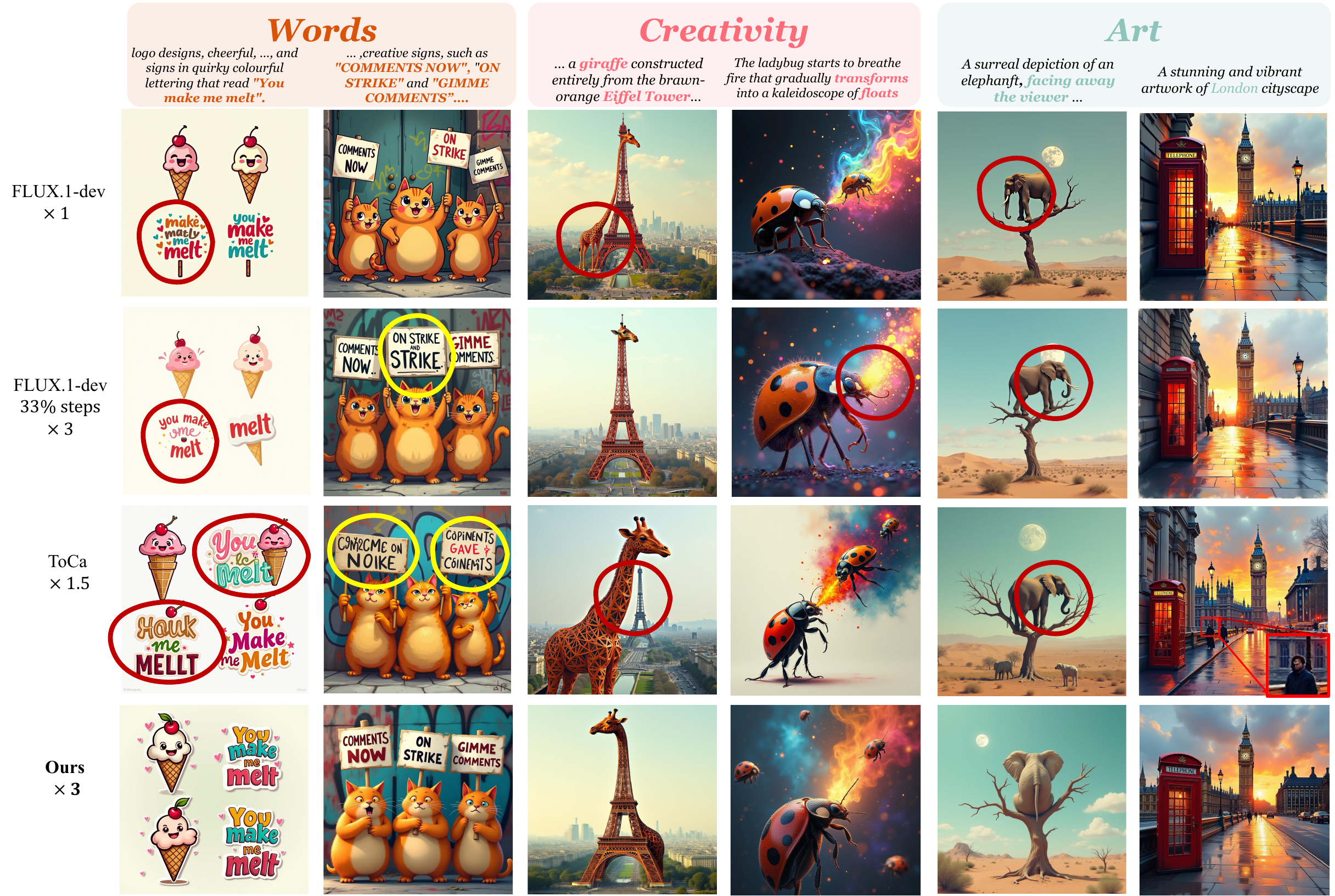}
    \caption{Qualitative comparison of our Bottleneck Sampling with FLUX.1-dev. Our method achieves up tp 3$\times$ speedup while maintaining or improving visual fidelity. Incorrect text rendering and anatomical inconsistencies are highlighted with different colors. \textit{Full prompts are provided in \cref{app:prompt_list}}.}
    \label{fig:exp-1}
\end{figure*}
We present the performance of Bottleneck Sampling in \cref{fig:exp-1} and \cref{tab:performance}. We compare it against the FLUX.1-dev baseline, which employs the FlowMatch scheduler with 50 inference steps, as well as accelerated FLUX variants that utilizes 33\% and 50\% of the original steps. Additionally, we include ToCa with 50 steps as a representative training-free acceleration method for further comparison. 

\paragraph{Qualitative Results} To evaluate robustness, we test each approach on challenging prompts that demand high fidelity and creativity, including text rendering, creative composition, and artistic generation. As shown in \cref{fig:exp-1}, Bottleneck Sampling consistently outperforms other methods in text rendering, accurately generating all letters without cherry-picking. In contrast, the baseline model occasionally misaligns characters, while cache-based methods like ToCa fail to produce coherent typography. We attribute this to partial attention map reuse, which may reduce effective computation needed for precise letter formation.
For creative and artistic prompts, Bottleneck Sampling maintains stronger semantic consistency. For instance, when generating descriptions like ``a giraffe constructed entirely from the Eiffel Tower" or ``an elephant facing away from the viewer," our method preserves coherence across resolution stages, reinforcing both structural integrity and textual fidelity. Moreover, it achieves threefold acceleration without quality degradation, maintaining critical details such as typography and semantic accuracy. These results demonstrate Bottleneck Sampling's effectiveness in enhancing efficiency while preserving high-quality generation.

\paragraph{Quantitative Results} As shown in \cref{tab:performance}, we compare the latency, FLOPs, and acceleration ratios of different models in the DiT computation process, along with their corresponding performance. To ensure a fair comparison, we exclude the computational cost of the text encoder and VAE, as these components remain identical across all models.
Across various evaluation metrics, Bottleneck Sampling consistently achieves superior performance at a baseline 2$\times$ acceleration, even outperforming ToCa’s 1.5$\times$ acceleration. Remarkably, our method maintains performance comparable to the baseline even at 3$\times$ acceleration. In more challenging benchmarks like T2I-CompBench, Bottleneck Sampling demonstrates a clear advantage, reinforcing findings from our controlled experiments.

\begin{table*}[ht]
\centering
\label{tab:performance}
\resizebox{0.95\linewidth}{!}{
\begin{tabular}{l|ccc|cccc}
\toprule
\textbf{Method} & \textbf{Latency(s)$\downarrow$} & \textbf{FLOPs(T)$\downarrow$} & \textbf{Speed $\uparrow$} &
\makecell{\textbf{CLIP} \\ \textbf{Score}$\uparrow$} &
\makecell{\textbf{Image} \\ \textbf{Reward}~\citep{xu2024imagereward} $\uparrow$} &
\makecell{\textbf{Gen} \\ \textbf{Eval}~\citep{xu2024imagereward} $\uparrow$} &
\makecell{\textbf{Average on T2I-} \\ \textbf{Compbench}$\uparrow$} 
\\ \midrule
FLUX.1-dev~\citep{flux} & 33.85 & 3719.50 & 1.00 $\times$ & 0.460 & 1.258 & 0.6807	 & 0.7032 \\ \midrule
33\% steps & 11.28 & 1239.83 & 3$\times$ & 0.432 & 1.048 & 0.6423 & 0.6140 \\ 
50\% steps & 16.93 & 1859.75 & 2$\times$ & 0.453 & 1.239 & 0.6698 & 0.6808 \\ 
ToCa~\citep{zou2024accelerating}  & 19.88 & 2458.06 & 1.51 $\times$ & 0.447 & 1.169 & 0.6630 & 0.6738 \\ 
\midrule
\rowcolor[gray]{0.9}\textbf{Bottleneck Sampling} ($\times$ 2) & 17.37 & 1870.10 & 2$\times$ & \textbf{0.460} & \textbf{1.257} & 0.6762 & 0.6820 \\ 
\rowcolor[gray]{0.9}  \textbf{Bottleneck Sampling} ($\times$ 3) & 14.46 &1234.39 & 3$\times$ & 0.457 & 1.254 & \textbf{0.6773} & \textbf{0.6946}\\ 
\bottomrule
\end{tabular}}
\caption{Quantitative Results on Text-to-Image Generation}

\end{table*}

\begin{table*}[ht]
\centering
\vspace{-1em}
\label{tab:performance2}
\resizebox{0.95\linewidth}{!}{
\begin{tabular}{l|ccc|ccc}
\toprule
\textbf{Method} & \textbf{Latency(s)$\downarrow$} & \textbf{FLOPs(T)$\downarrow$} & \textbf{Speed $\uparrow$} & \textbf{CLIP Score $\uparrow$} & \textbf{Vbench $\uparrow$} & \textbf{T2V-Compbench $\uparrow$} \\ \midrule
HunyuanVideo~\citep{kong2025hunyuanvideosystematicframeworklarge} & 1896 & 605459.49 & 1.00$\times$ &  0.455 & 83.24 & 0.5832 \\ \midrule
40\% steps & 758.4 & 242183.80& 2.5$\times$  & 0.427 & 82.93 & 0.5530 \\ 
50\% steps & 948.0 & 302729.74 & 2$\times$ & 0.439 & 83.14 & 0.5631 \\ 
\midrule
\textbf{Bottleneck Sampling} ($\times$ 2) & 1321.4  & 311319.54& 2$\times$  & \textbf{0.446} & 83.18 & \textbf{0.5739} \\ 
\rowcolor[gray]{0.9}\textbf{Bottleneck Sampling} ($\times$ 2.5) &834.7  &  232756.28 & 2.5$\times$ & 0.443 & \textbf{83.19} & 0.5737 \\ 
\textbf{Bottleneck Sampling} ($\times$ 3) & 743.2 & 203192.35 & 3$\times$  & 0.421 & 81.98 & 0.5626 \\ 
\bottomrule
\end{tabular}}
\caption{Quantitative Results on Text-to-Video Generation}
\end{table*}

\subsection{Results on Video Generation}  
\begin{figure*}[ht]
    \centering
    \includegraphics[width=1\linewidth]{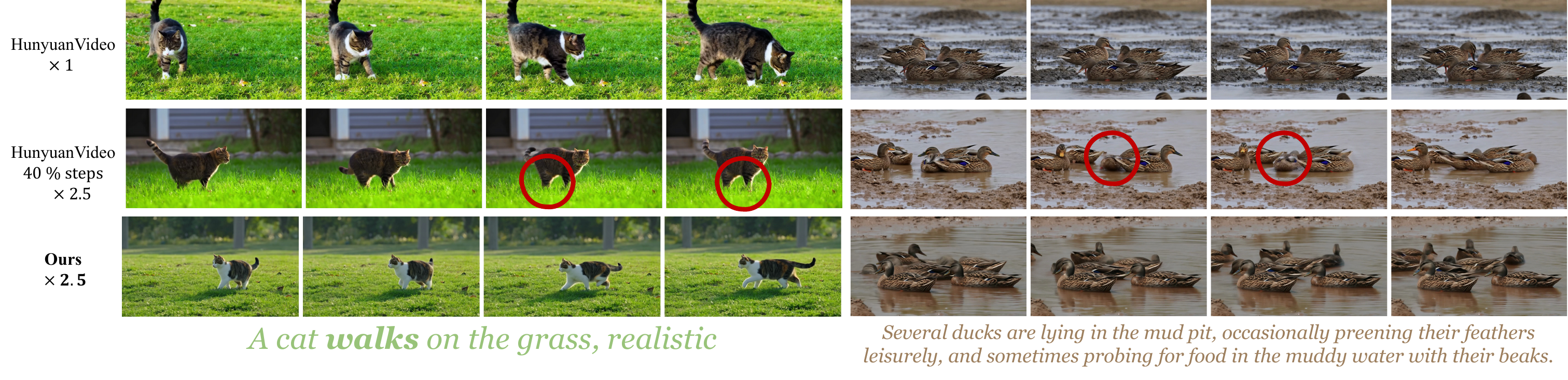}
    \vspace{-2em}
    \caption{Qualitative comparison of our Bottleneck Sampling with HunyuanVideo~\citep{kong2025hunyuanvideosystematicframeworklarge}.Our method achieves up to 2.5$\times$ speedup while maintaining visual fidelity. Incorrect object motion and object disappearing are highlighted with red circles.}
    \label{fig:exp-2}
    \vspace{-1.1em}
\end{figure*}

We present the results of our video generation experiments in \cref{fig:exp-2} and \cref{tab:performance2}. The baseline model is HunyuanVideo~\cite{kong2025hunyuanvideosystematicframeworklarge} with 50 inference steps. Due to the lack of open-source implementations based on the HunyuanVideo architecture, we compare our method against an equivalent acceleration variant of the original model with a reduced number of inference steps. ToCa's reported results are based on Open-Sora~\citep{opensora}, an outdated and weaker baseline that lacks modern generative architectures such as MM-DiT and flow matching. To ensure a fair comparison, we exclude ToCa from our evaluation.

As shown in \cref{fig:exp-2}, Bottleneck Sampling effectively preserves motion details, background consistency, semantic coherence, and composition fidelity, maintaining a quality level comparable to the baseline. In contrast, the reduced-step baseline introduces noticeable artifacts and inconsistencies, often struggling with semantic misalignment (e.g., abrupt motion stoppages, such as a cat suddenly halting) and visual defects (e.g., object disappearance or incorrect motion trajectories, like a duck vanishing mid-frame). Additional qualitative results are provided in \cref{app:more}.
Furthermore, as reported in \cref{tab:performance2}, Bottleneck Sampling achieves performance on par with the baseline across multiple evaluation metrics, including VBench, T2V-CompBench, and CLIP Score, while attaining up to 2.5× acceleration. These results further highlight the efficiency of our approach in accelerating video generation without compromising output quality.

\subsection{Ablation Study}  

\label{ablation}

\begin{figure}[ht]
\vspace{-0.5em}
    \centering
    \includegraphics[width=1\linewidth]{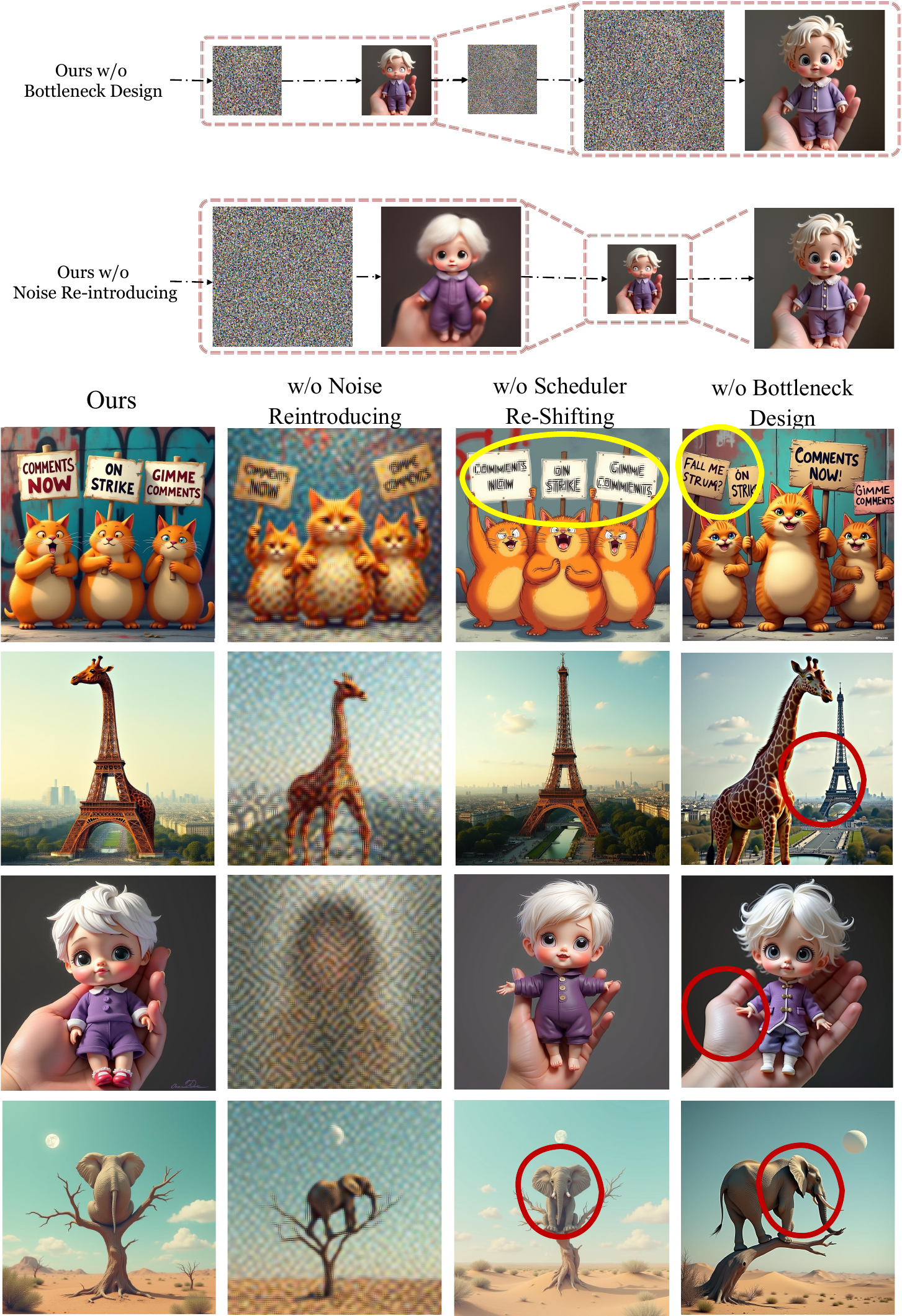}
    \caption{Visualization of Ablation Studies.}
    \label{fig:ablation}
    \vspace{-1.8em}
\end{figure}


In this section, we investigate the key design choices in Bottleneck Sampling through ablation studies, including the overall bottleneck design, noise reintroducing at stage transitions, and scheduler re-shifting. We report quantitative results for image generation in \cref{tab:ablation} and provide qualitative evaluations in \cref{fig:ablation}. Additional ablation studies, including upsampling methods and stage configurations, are provided in \cref{app:ablation}.

\paragraph{Effect of Noise Reintroducing}
We examine the impact of noise reintroducing at stage change points. In a standard cascaded diffusion setup, noise remains continuous without intermediate reintroduction. However, results in \cref{fig:ablation} indicate that this leads to persistent ghosting artifacts and overlapping pixels in high-resolution inference, severely degrading quality. By contrast, our noise reintroduction strategy alleviates these issues, enabling cleaner and more coherent outputs.  

\paragraph{Effect of Scheduler Re-Shifting}
To assess the impact of scheduler re-shifting, we conduct an ablation study where no re-shifting is applied and a fixed scheduling is applied. As shown in \cref{fig:ablation}, this results in detail loss and increased ghosting, consistent with our prior observations. These artifacts arise from signal-to-noise ratio variations across stages, where a static schedule leads to suboptimal denoising. In contrast, our re-shifting mechanism mitigates these issues, ensuring a more stable and high-quality generation.

\paragraph{Effect of Bottleneck Design}  
Another common approach to leveraging low-resolution priors is the cascaded diffusion framework~\citep{ho2022cascaded, jin2024pyramidal}, where denoising is first performed at a lower resolution before upsampling and refinement. We conduct comparative experiments, ensuring that Bottleneck Sampling and cascaded-like sampling operate under the \textbf{same acceleration ratio}, as shown in \cref{fig:ablation}. The results demonstrate that our design, which introduces an initial high-resolution stage, consistently improves semantic coherence, particularly in text rendering and fine-detail preservation and achieves better generation quality.  

\begin{table}[ht]
    \centering
    \label{tab:ablation}
    \resizebox{0.98\linewidth}{!}{
    \begin{tabular}{lcc}
        \toprule
        Method & CLIP Score & ImageReward \\
        \midrule
        Baseline & 0.460 & 1.258 \\
        Ours w/o Scheduler Re-Shifting & 0.276 & 0.781 \\
        Ours w/o Noise Reintroducing & 0.379 & 0.923 \\
        Ours w/o Bottleneck Design & 0.448 & 1.134\\
        \textbf{Ours $\times$ 3} & \textbf{0.460} & \textbf{1.257} \\
        \bottomrule
    \end{tabular}}
    \caption{Ablation Study Results on Image Generation.}
    \vspace{-1.5em}
\end{table}



\section{Conclusion}
Motivated by the observation that diffusion models are pre-trained on diverse image and video resolutions, we introduce \textbf{Bottleneck Sampling}, a training-free method that accelerates inference by strategically leveraging low-resolution priors. Following a high-low-high workflow, Bottleneck Sampling reduces computation by primarily performing denoising at lower resolutions in intermediate steps while refining details at high resolutions at the start and end. Our method achieves 3$\times$ speedup for image generation and 2.5$\times$ for video generation, surpassing previous training-free acceleration approaches while maintaining output generation quality. These findings provide insights into future training-free acceleration in diffusion models.

{
    \small
    \bibliographystyle{ieeenat_fullname}
    \bibliography{main}
}
\clearpage
\setcounter{page}{1}
\maketitlesupplementary
\appendix

\section{Implementaion Details}

\subsection{Baseline Models Configuration}
In this section, we describe the configurations of different baseline models used in our study. We adopt the original model implementations whenever possible. 

\begin{itemize}
    \item \textbf{FLUX.1-dev~\citep{flux}:} A 12-billion-parameter rectified flow transformer designed for text-to-image generation. Built upon the MMDiT architecture\citep{esser2024scaling}, FLUX.1-dev scales to 12B parameters and consistently achieves state-of-the-art performance in image generation. However, its computational cost remains substantial, requiring up to 30 seconds on A100 80G to generate one 1024p image.
    
    \item \textbf{HunyuanVideo~\citep{kong2025hunyuanvideosystematicframeworklarge}:} A 13-billion-parameter open-source video generation model. HunyuanVideo is recognized for its smooth motion synthesis, precise semantic alignment, and high-quality aesthetics. However, its computational demands are considerable, requiring 50 minutes on an A100 80G or 30 minutes on an H100 80G to generate a 1280p video with 129 frames, posing significant challenges for practical deployment.
\end{itemize}

\subsection{Evaluation Metrics}
For text-to-image generation evaluation, we curated a set of 400 prompts, manually selected from HPSv2~\citep{hps} and high-quality web-sourced prompts, as detailed in~\cref{app:prompt_list}. These prompts were chosen to emphasize complex, fine-grained, and detail-rich scenarios, often involving multiple objects. The generated images were evaluated using several widely adopted benchmarks, including CLIP~\citep{radford2021learning}, which assesses text-image alignment, ImageReward~\citep{xu2024imagereward}, a reward model trained to reflect human preferences, GenEval~\citep{ghosh2023geneval}, and T2I-CompBench~\citep{huang2023t2i}, which measure compositional abilities. For GenEval and T2I-CompBench, we report the average scores across all metrics.

For text-to-video generation evaluation, we mannualy selected a subset of prompts from VBench~\citep{huang2024vbench}, T2V-CompBench~\citep{sun2024t2v},  and Hunyuan Evaluation Prompts~\citep{kong2025hunyuanvideosystematicframeworklarge}, focusing on longer and more descriptive text inputs to better assess complex generative capabilities. Each method generated five videos per prompt using different random seeds to ensure a more comprehensive comparison. The generated videos were evaluated using per-frame averaged CLIP Score~\citep{radford2021learning}, the Compositional Video Benchmark~\citep{sun2024t2v}, and the widely used VBench~\cite{huang2024vbench}, which categorizes performance into two primary scores: quality and semantic accuracy. To reflect the greater importance of visual fidelity in generative evaluation, a weighted total score was computed, assigning the quality dimension four times the weight of the semantic dimension.

\subsection{Hyperparameter}
\label{app:hyp}
In this section, we further detailed our selected hyperparameters for bottleneck sampling on both text-to-image generation and text-to-video generation in our presented results. 
\begin{table}[h!]
    \centering
    
    \resizebox{0.98\linewidth}{!}{
    
    \begin{tabular}{l|c}
        \toprule
        \textbf{Baseline} &\\ 
        height  $h$ & $1024$\\
        width  $w$ & $1024$ \\
        number of inference steps $N$&  $50$  \\
        shifting $s$& $3.0$ \\
        \midrule
        \textbf{Bottleneck Sampling} &\\ 
        stage $K$ & 3 \\
        height list $\{h_i\}_{i=0}^K$ & $[1024,512,1024]$\\
        width list $\{w_i\}_{i=0}^K$ & $[1024,512,1024]$ \\
        \midrule
        \textbf{Noise Reintroducing} & \\
        strength list $\{w_i\}_{i=0}^K$& $[1,0.8,0.6]$ \\
        number of inference steps $\{N_i\}_{i=0}^K$&  $[6,20,8]$  \\
        \midrule
        \textbf{Scheduler Re-Shifting} & \\
        shifting list $\{s_i\}_{i=0}^K$& $[9,6,9]$ \\
        \bottomrule
    \end{tabular}}
    \caption{
        Hyperparameters of Bottleneck Sampling for Text-to-Image Generation on FLUX.1-dev~\citep{flux}.
    }
    
    \label{tab:hyperparameters}
\end{table}

\begin{table}[h!]
    \centering
    
    \resizebox{0.98\linewidth}{!}{
    
    \begin{tabular}{l|c}
        \toprule
        \textbf{Baseline} &\\ 
        height  $h$ & $720$\\
        width  $w$ & $1280$ \\
        number of inference steps $N$&  $50$  \\
        shifting $s$& $7.0$ \\
        \midrule
        \textbf{Bottleneck Sampling} &\\ 
        \textbf{Stage Configuration} &\\ 
        stage $K$ & 3 \\
        height list $\{h_i\}_{i=0}^K$ & $[720, 544, 720]$\\
        width list $\{w_i\}_{i=0}^K$ & $[1280, 960, 1280]$ \\
        frames list  $\{t_i\}_{i=0}^K$ & $[129, 129, 129]$ \\
        \midrule
        \textbf{Noise Reintroducing} & \\
        strength list $\{w_i\}_{i=0}^K$& $[1,0.8,0.7]$ \\
        number of inference steps $\{N_i\}_{i=0}^K$&  $[4,24,16]$  \\
        \midrule
        \textbf{Scheduler Re-Shifting} & \\
        shifting list $\{s_i\}_{i=0}^K$& $[7,9,9]$ \\
        \bottomrule
    \end{tabular}}
    \caption{
        Hyperparameters of Bottleneck Sampling for Text-to-Video Generation on HunyuanVideo~\citep{kong2025hunyuanvideosystematicframeworklarge}.
    }
    \label{tab:hyperparameters2}
\end{table}

\section{Computational Complexity Analysis}

In this section, we conduct a detailed complexity analysis to quantify the computational savings achieved by Bottleneck Sampling.
Taking an image generation model as an example, given a target resolution \( H \times W \), the DiT model first compresses the input using a VAE and then applies a patchification operation before computing attention. In the case of FLUX, the VAE compression ratio is \( 8 \), and the patch size is \( 2 \), leading to an attention sequence length of $S = \frac{H}{16} \times \frac{W}{16}.$ Since MMDiT processes both text and image tokens together in a unified self-attention mechanism, and the text sequence length is typically an order of magnitude smaller than the image sequence length, we approximate the total sequence length as:$\frac{H}{16} \times \frac{W}{16}$

During computation, tokens are first projected through a linear layer to obtain queries, keys, and values. The queries and keys are then multiplied via a dot product, passed through a softmax function, and multiplied with the values. Finally, the output undergoes another linear transformation. The total computational cost of the Self-Attention layer can be expressed as:

\begin{align}
    \text{FLOPs} &= 6 \times S \times D^2 + 4 \times S^2 \times D + 2 \times S^2  \times n \notag \\ 
    &+ 2 \times S \times D^2 + 16 \times S \times D^2.
\end{align}

where $S$ denotes the sequence length, $D$ denotes the hidden dim, and $n$ denotes the number of heads. 
Simplifying and neglect the number of heads which is relatively small, we obtain:

\begin{align}
    \text{FLOPs} \approx 24 \times S \times D^2 + 4 \times S^2 \times D
\end{align}

This expression provides a clear measure of the computational cost associated with self-attention in diffusion transformers.

Taking into account real values, where \( D = 3072 \), we compute:
\[
\text{FLOPs}_{512} \approx 20.43 \text{ TFlops}, \quad \text{FLOPs}_{1024} \approx 78.86 \text{ TFlops}.
\]

Thus, at each step of computation (NFE), reducing the resolution by half results in a speedup of:

\begin{align}
    \frac{\text{FLOPs}_{1024}}{\text{FLOPs}_{512}} \approx 3.8.
\end{align}

In summary, leveraging lower-resolution inference during intermediate steps, where fine-grained details are less critical, significantly reduces computational FLOPs while maintaining comparable performance.






\section{User Study}
\label{app:user_study}

Given the inherent randomness and diversity in image and video generation, relying solely on quantitative metrics may be insufficient for a comprehensive evaluation of performance. To further validate our results, we conducted a user study to assess the perceptual quality of generated outputs.

We randomly sampled 100 image prompts and 100 video prompts and recruited five trained human annotators for evaluation. For each prompt, a guidance scale was randomly selected from three broad ranges during inference. We then generated one pair of outputs: one from the baseline model (FLUX/1-dev for images and HunyuanVideo for videos) and the other from Bottleneck Sampling. The annotators were presented with these outputs in an anonymized manner and asked to select one of three options: Model 1, Model 2, or Same. This process ensured an unbiased assessment of visual quality and text alignment.

The results of the user study for both text-to-image and text-to-video tasks are summarized in \cref{tab:user_study}.

\begin{table}[ht]
\centering

\resizebox{0.98\linewidth}{!}{
\begin{tabular}{l|ccc}
\toprule
Task & Baseline & Ours & Same  \\ \midrule
FLUX.1-dev~\citep{flux} & 16\% & 17\% & 67\% \\
HunyuanVideo~\cite{kong2025hunyuanvideosystematicframeworklarge} & 14\% & 13\% & 73\% \\ \bottomrule
\end{tabular}}
\caption{User study results for text-to-image and text-to-video tasks, measured by the selection rate for each option.}
\label{tab:user_study}
\end{table}

The study results indicate that in most cases, the outputs from our Bottleneck Sampling model are perceptually comparable to those of the baseline models, with the majority of annotators selecting the ``Same" option. This suggests that our approach preserves high-quality generation while significantly improving efficiency, making it a viable alternative for computationally constrained settings.

\section{Extended Ablation Studies}
\label{app:ablation}

In this section, we conduct additional ablation studies to evaluate the robustness of our approach.

\begin{table}[ht]
\centering
\resizebox{0.98\linewidth}{!}{
\begin{tabular}{l|cc}
\toprule
Sampling Stages & CLIP Score $\uparrow$ & ImageReward $\uparrow$ \\ 
5 Stage & 0.451 & 1.253  \\ 
\textbf{3 Stage} & \textbf{0.457} & \textbf{1.254} \\
\midrule
Upsampling Method & CLIP Score $\uparrow$ & ImageReward $\uparrow$ \\ \midrule
Bilinear Interpolation & 0.453 & 1.249 \\
Bicubic Interpolation & 0.454 & 1.251 \\
Nearest-Neighbor Interpolation & \textbf{0.461} & 1.249 \\
\textbf{Lanczos Interpolation} & 0.457 & \textbf{1.254} \\ \bottomrule
\end{tabular}}
\caption{Comparison of different upsampling methods based on CLIP Score and ImageReward.}
\label{tab:upsampling_comparison}
\end{table}

\subsection{Effect of Stage Numbers}
\begin{figure}[ht]
    \centering
    \includegraphics[width=1\linewidth]{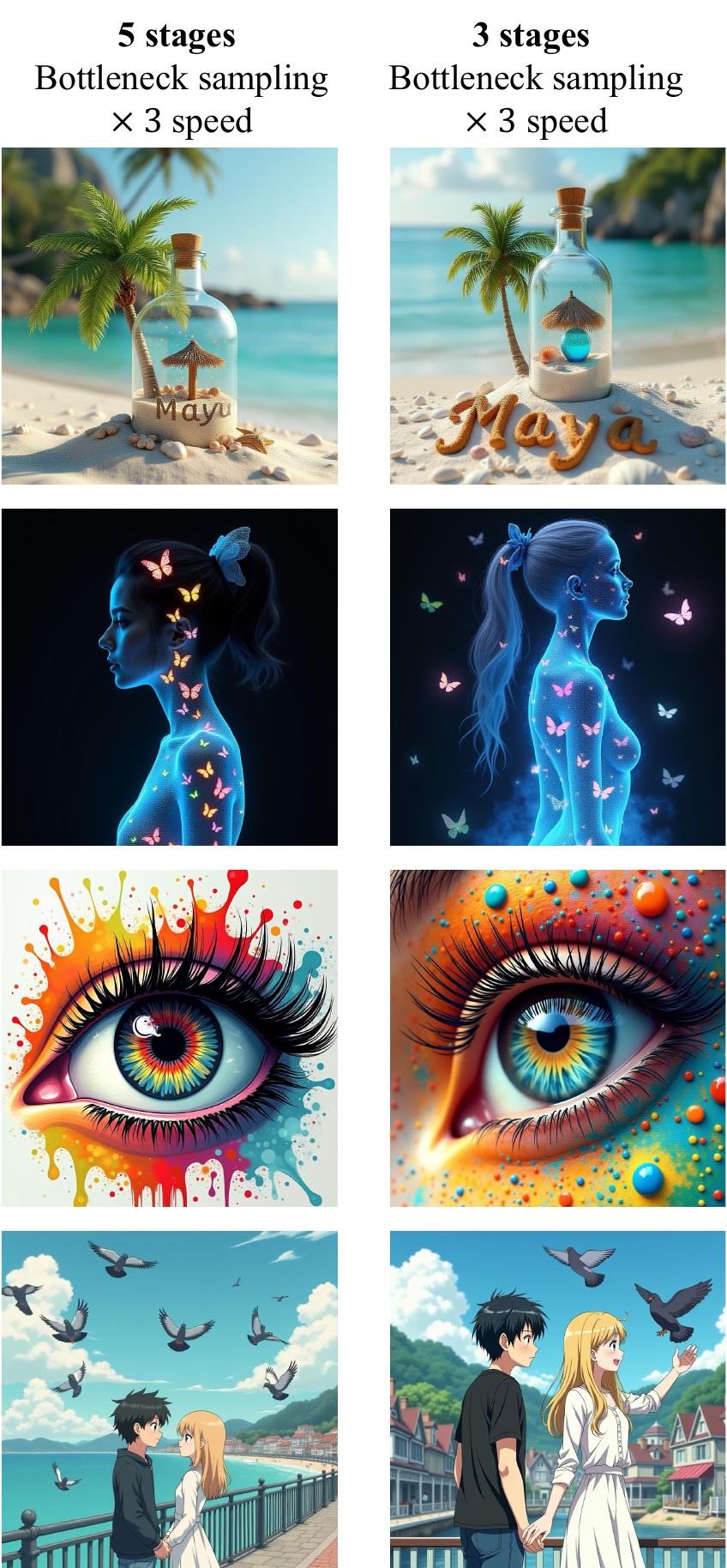}
    \caption{Effect of Stage Numbers in our Bottleneck Sampling. Comparison between 3 stages and 5 stages. }
    \label{fig:ab3}
\end{figure}

Our primary experiments report results based on a three-stage configuration. To further investigate the impact of stage design, we conduct additional ablations by increasing the number of stages. In \cref{fig:ab3} and \cref{tab:upsampling_comparison}, we present results on FLUX.1-dev using a five-stage setting: \([1024, 512, 256, 512, 1024]\). The performance remains comparable to the three-stage configuration, with the latter potentially offering a more favorable trade-off.

Although a five-stage Bottleneck Sampling setup may theoretically provide a higher upper bound in performance, it introduces significantly more hyperparameters. Due to the lack of a direct self-evaluation mechanism in diffusion models, determining these hyperparameters can be challenging. As such, we adopt the three-stage Bottleneck Sampling configuration for this study and leave further exploration of multi-stage designs to future work.

\subsection{Effect of Upsampling Methods}

\begin{figure}[ht]
    \centering
    \includegraphics[width=1\linewidth]{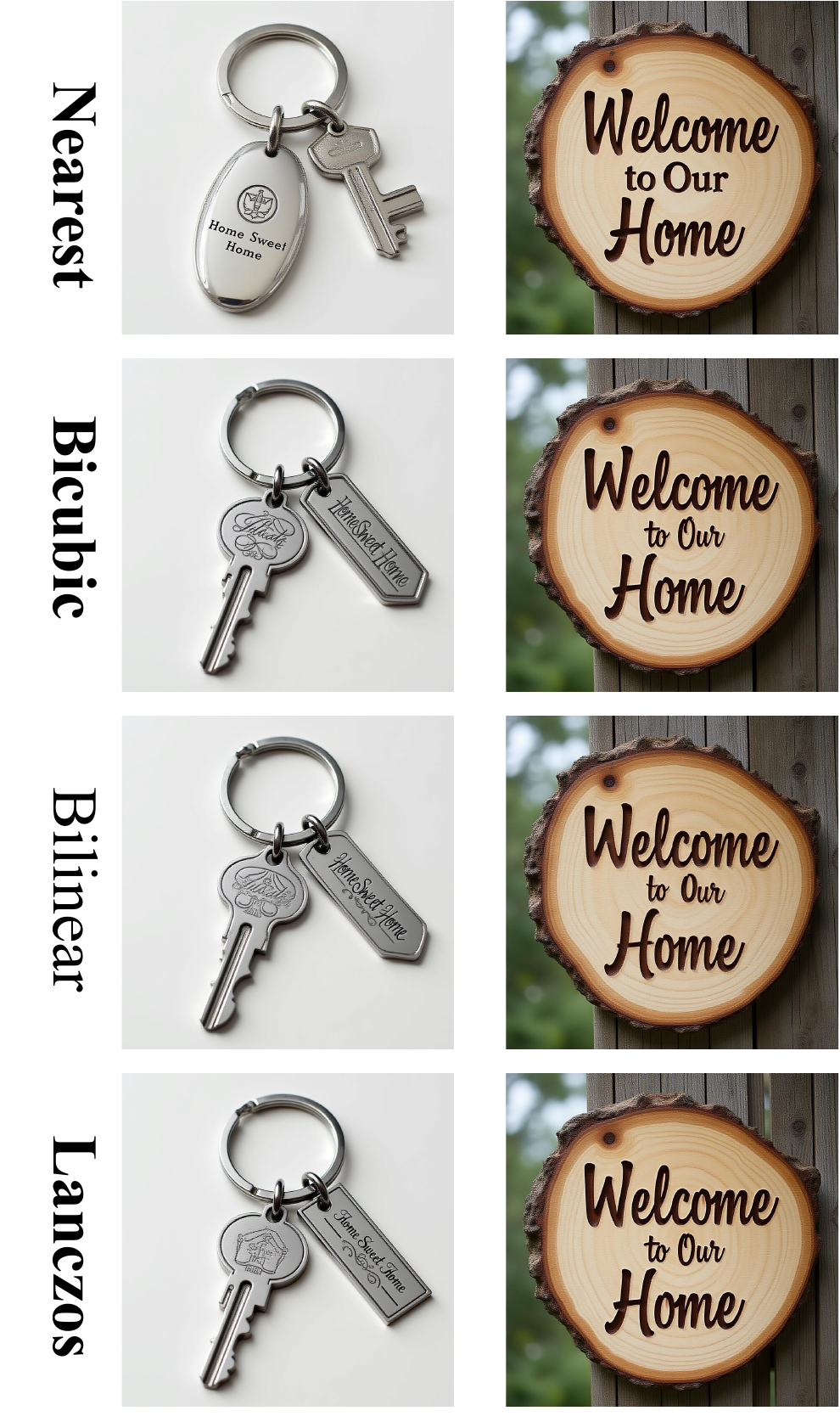}
    \caption{Effect of Upsampling Method in our Bottleneck Sampling. }
    \label{fig:ab4}
\end{figure}

We evaluate the impact of different upsampling methods on performance. For simplicity and ease of implementation, our default choice is bilinear interpolation. However, to explore alternative upsampling strategies, we conduct experiments with the following commonly used methods:

\begin{enumerate}
    \item \textbf{Bilinear Interpolation}: Computes the output pixel as a weighted average of the four nearest input pixels, offering a balance between speed and smoothness.
    \item \textbf{Bicubic Interpolation~\citep{keys1981cubic}}: Uses a cubic convolution to estimate pixel values based on the 16 nearest neighbors, typically producing smoother results than bilinear interpolation.
    \item \textbf{Nearest-Neighbor Interpolation}: Assigns each output pixel the value of its closest input pixel, preserving sharp edges but often introducing blocky artifacts.
    \item \textbf{Lanczos Interpolation~\citep{duchon1979lanczos}}: A high-quality resampling method using sinc functions to achieve sharper results, at the cost of higher computational overhead.
\end{enumerate}

We report the impact of different upsampling methods on model performance in \cref{tab:upsampling_comparison} and \cref{fig:ab4}.
The results indicate that upsampling methods have a minimal effect on overall generation quality, with most methods performing comparably across different images. For text-based image generation, nearest-neighbor interpolation achieves the highest text accuracy but may introduce inconsistencies in visual style. To further assess human perceptual preferences, we evaluate ImageReward scores and ultimately adopt Lanczos interpolation in our final setup.
Notably, all upsampling methods produce satisfactory results within our pipeline, highlighting the robustness of the proposed approach.


\section{Prompt list}
\label{app:prompt_list}
We provide our prompt list for the generated images presented in our qualitative results and part of our evaluation prompt sets as follows:

\begin{enumerate}
    \item Brunette pilot girl in a snowstorm, full body, moody lighting, intricate details, depth of field, outdoors, Fujifilm XT3, RAW, 8K UHD, film grain, Unreal Engine 5, ray tracing.
\item An ancient spiritual gnomes stone pathway rock garden.sculptured . Style of alex grey,giger.unreal engine.totem sculptures.swirling patterned stone pathway courtyard.wood.stone.driftwood.statues. artistic sculpture.lanterns.air bnb tiny house
\item A droplet from a small brook enters the ocean, creating rippling water. A few bamboo leaves, white and green, are seen with a soft focus effect, glinting under the sun's twinkling rays.
\item A blue coloured pizza.
\item A new Human Mecha combined, set against a massive post-apocalyptic background, realisticlighting, ruined ruins, high level of rendering, virtual reality.
\item A racing car with a silver transparent texture, showcasing design sensibility against a white background, industry design.
\item A beautiful woman facing to the camera, smiling confidently, colorful long hair, diamond necklace, deep red lip, medium shot, highly detailed, realistic, masterpiece.
\item Realism, Unreal Engine, cinematic feel, exaggerated lighting, cyberpunk, future world, advanced technology, neon city at night, a car chase scene, a busy road with many cars coming and going, a police car is chasing a yellow taxi, with a strong sense of speed, exaggerated lens effects, and movie screenshots.
\item  Japanese anime, celluloid style, animation screenshots, cyberpunk, future world, technologically advanced, Close-up, A capable woman with dark blue short hair, wearing high-tech outfit, is driving, with a nervous expression.
\item American cartoons, 3D modeling, commercial cartoon movies, high-definition rendering, cyberpunk, future world, advanced technology, black interior environment, close-up, backlight, black and blue tones, cold atmosphere, machinery factory. There is a silver robot head suspended by cables.
\item Realism, Unreal Engine, cinematic feel, exaggerated lighting, horror movie, over-the-shoulder shot, the center of the picture shows an ancient seaside lighthouse, shrouded in thick fog, exuding a gloomy atmosphere, the close shot on the left side of the picture is a woman in her twenties with medium-length black hair sitting on a boat looking at the lighthouse, she is wearing a black shirt, a brown suit jacket and long jeans.
\item Japanese anime, celluloid style, animation screenshots,Medium shot, with a charming American seaside town in the background, a boy with short black hair, in his twenties, wearing a black shirt and jeans, and a girl with medium-long blond hair, in her twenties, wearing a white long-sleeved dress, standing and feeding the pigeons together.
\item 2D cartoon,Diagonal composition, Medium close-up, a whole body of a classical doll being held by a hand, the doll of a young boy with white hair dressed in purple, He has pale skin and white eyes.
\item An enchanting dark fantasy scene captures a semi-transparent woman very perfect mind formed woman body,ponytail hair and a bow on the side of the head with her silhouette illuminated by a radiant blue hue, accentuating her graceful form. Surrounded by a mesmerizing array of glow-in-the-dark butterflies in vibrant neon colors like electric green, pulsating pink, and luminous orange, she seems to be the center of a surreal spectacle of radiant beauty the silhouette has a reflection as if it were made of glass and shines as if it had varnish. The butterflies, varying in size, fill the silhouette completely, creating a breathtaking visual experience. The clean darkness of the background serves as a perfect contrast, evoking a sense of enchantment, wonder, and mystique. This conceptual artwork masterfully combines elements of wildlife photography, cinematic aesthetics, and fashion illustration to create a dreamlike realm where magic and mystery intertwine. wildlife photography, fashion, conceptual art, ukiyo-e, 3d render, cinematic, photo, dark fantasy, illustration, vibrant, portrait photography
\item A vibrant and colorful artistic representation of an eye. The eye is the central focus, with its iris exhibiting a spectrum of colors ranging from reds to blues. Surrounding the eye are abstract patterns and shapes in a myriad of colors, including oranges, yellows, greens, and blues. Some of these patterns resemble fluid or paint splatters, while others have a more structured, almost psychedelic appearance. The eye's eyelashes are prominently depicted, and they seem to be made of thick, black strands. The overall feel of the image is dynamic, energetic, and evocative of intense emotion or creativity.
\item An awe-inspiring 3D render of a glass bottle magically transformed into a miniature tropical paradise. The pristine white sand glistens beneath the warm sunlight, accompanied by sparkling shells and an intricately detailed palm tree swaying gently. A charming thatched-roof hut and a shimmering blue bottle add to the idyllic beach setting, while the sandy message ``Maya "exudes joy and tranquility. The cinematic and illustrative style of the rendering immerses the viewer in a sun-drenched, captivating escape. This 32k, 4D, full HDR, and hyper-realistic masterpiece is a testament to the future of digital art and imaging, redefining visual storytelling through its breathtaking depth and detail., poster, typography, illustration, photo, 3d render, cinematic
\item A whimsical and creative digital art of a giraffe constructed entirely from the brawn-orange Eiffel Tower. Each metal tower segment bends to form the long neck and body of the giraffe, with its iconic pointed tip serving as the head. The background reveals a bustling cityscape, where the Arc de Triomphe and other Parisian landmarks emerge in the distance. The overall atmosphere of the digital illustration is playful, offering a unique and unexpected perspective on a world-famous monument. conceptual art
\item A surreal depiction of an elephant, facing away the viewer, sitting on a thin, fragile looking bare tree branch, set against a serene desert landscape under a clear sky. Big elephant sitting on a tree branch in the desert with a moon in the background, surrealistic digital artwork, surrealism 8k, 4 k surrealism, Below the big elephant is a vast desert with sand mounds and sparse vegetation. Clear blue sky above, African landscape, overcast, candid photo, cinematic, countryside, curious animals, English, sunset, beautiful landscape, vibrant sky, (((greenery))), (((wide shot))), wildlife, hyper realistic photo, 8k, high resolution, (((lambs))), clear faces, lamb fur, realistic, excellent light, shadows. Super details, illustration, 3d render, painting
\item A captivating conceptual art piece featuring a ladybug as the centerpiece. The ladybug, initially depicted in monochrome, starts to breathe fire that gradually transforms into a kaleidoscope of vibrant colors where other ladybug-suited big roach gracefully floats among a breathtaking display of vibrant intergalactic elements. The background is minimal and largely blank, allowing the viewer's focus to be entirely on the ladybug and its mesmerizing transformation. The artwork conveys a sense of rebirth and the power of change through its striking contrast of color palettes., conceptual art
\item A stunning and vibrant artwork of London cityscape, showcasing the iconic Big Ben clock tower as the focal point. The tower stands tall and majestic, with a glowing orange and yellow sunset casting a warm glow over the scene. To the left, a classic red telephone booth adds a touch of traditional British charm, its reflection mirrored in the wet, glistening streets. The streets have a dreamy, watercolor-like quality, with muted grays, vibrant reds, and splashes of blue in the reflections of the buildings and sky. The overall composition is reminiscent of a movie scene, capturing the essence of London in a captivating and artistic way., photo, cinematic, poster, vibrant, painting, illustration, portrait photography
\item Imagine a rustic wooden sign for a bed and breakfast, featuring some elegant welcoming words ``Welcome to Our Home" carved into the wood,  in a flowing script that complements the natural grain of the wood.
\item A chic keychain with a small, elegant logo engraved on one side and the words ``Home Sweet Home" written underneath, in a polished metal finish
\item Design a stylish dancer's back logo with the letters ``R" and ``Y"
\item A humorous and vibrant digital art piece featuring a group of chubby, orange cats serving as protesters. They are holding up witty and creative signs, such as ``COMMENTS NOW", ``ON STRIKE" and ``GIMME COMMENTS". The cats have expressions of determination and urgency, standing in front of a graffiti-covered wall. The overall atmosphere is lively and humorous, with a touch of rebellion
\item logo designs, cheerful cartoon ice cream cones topped with a cherry and a bright smile, and signs in quirky colourful lettering that read ``You make me melt".
\end{enumerate}

\section{More Qualitative Results}
\label{app:more}
\begin{figure*}[ht]
    \centering
    \includegraphics[width=1\linewidth]{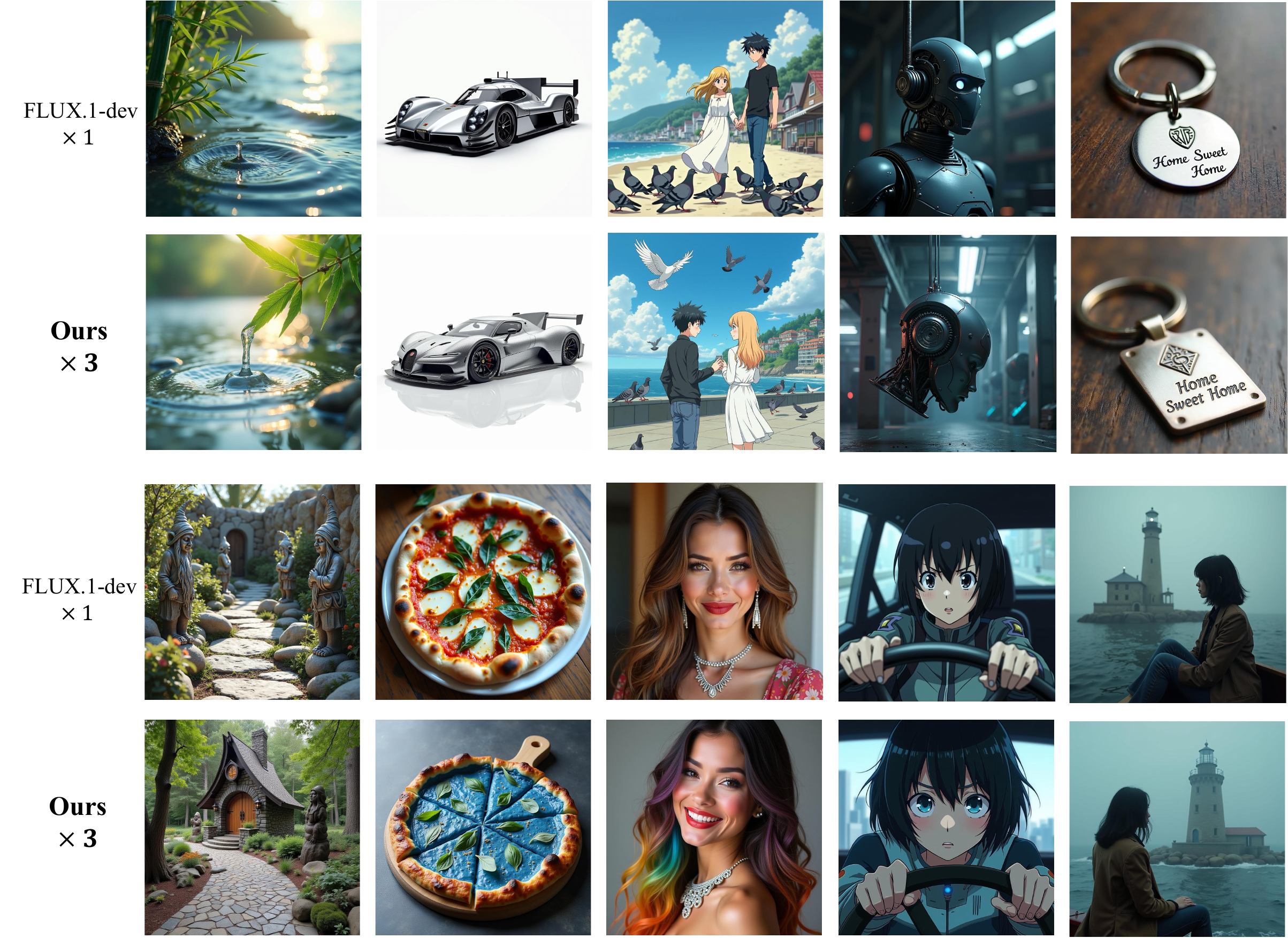}
    \caption{Comparison with FLUX.1-dev}
    \label{fig:append1}
\end{figure*}

\begin{figure*}[ht]
    \centering
    \includegraphics[width=1\linewidth]{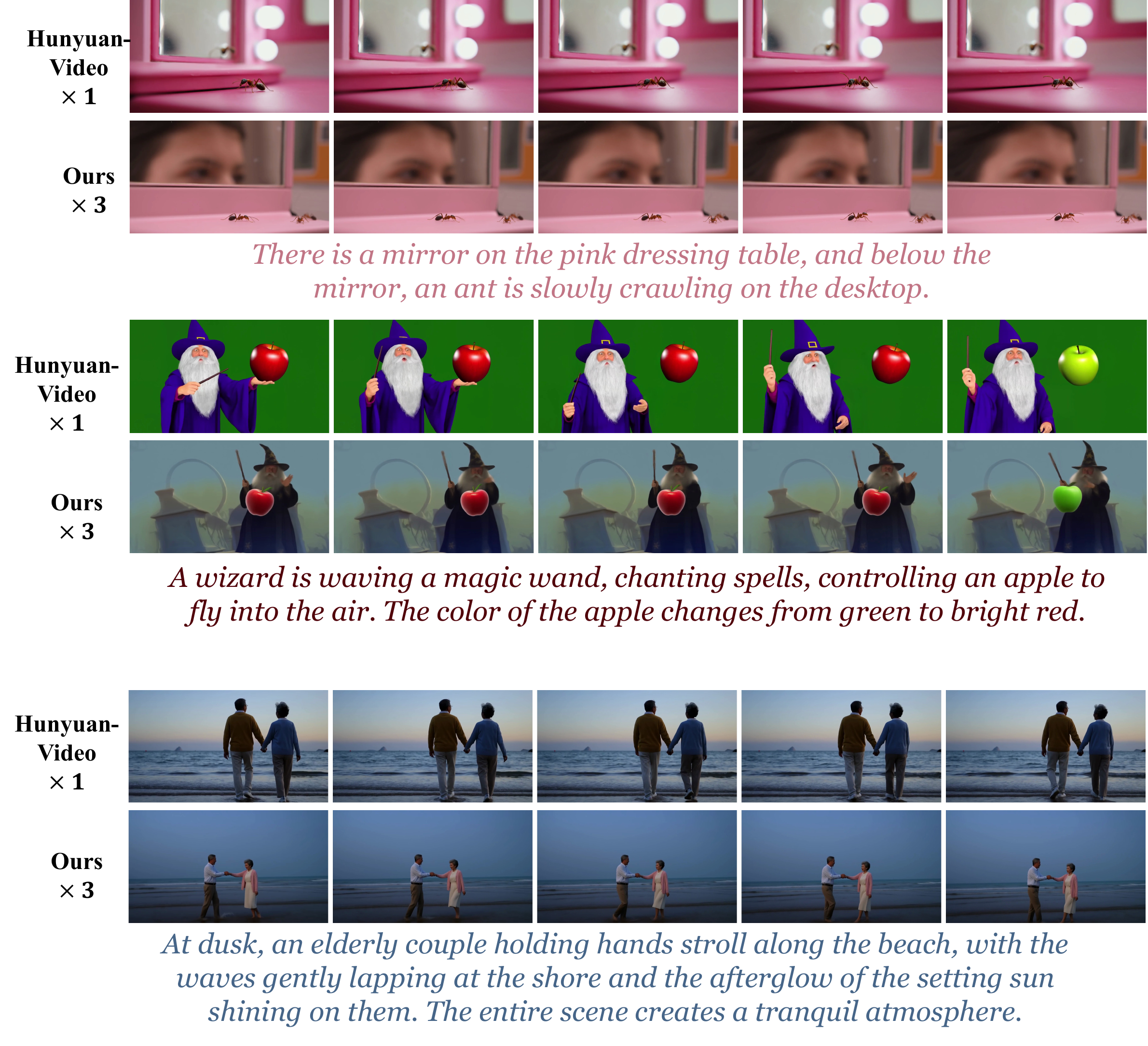}
    \caption{Comparison with HunyuanVideo}
    \label{fig:append2}
\end{figure*}

We provide a detailed comparison with baseline models in \cref{fig:append1} and \cref{fig:append2}.

\end{document}